\documentclass[a4paper, 10pt, conference]{ieeeconf}      

\usepackage[backend=biber,style=ieee,sorting=none]{biblatex}
\DeclareCiteCommand{\autocite}%
{\usebibmacro{prenote}\bibopenbracket}%
{\usebibmacro{citeindex}%
	\printfield{labelnumber}}%
{\multicitedelim}%
{\bibclosebracket\usebibmacro{postnote}}

\usepackage{graphicx}
\usepackage{float}
\usepackage{subfigure}
\usepackage{blindtext}
\usepackage{lscape}
\usepackage{rotating}
\usepackage{tikz}
\usepackage{pgfplots}
\pgfplotsset{compat=1.18}

\newcommand\IEEEhyperrefsetup{
	bookmarks=true,bookmarksnumbered=true,%
	colorlinks=true,linkcolor={black},citecolor={black},urlcolor={black}%
}
\usepackage[\IEEEhyperrefsetup, pdftex]{hyperref}

\usepackage{titlesec}
\titlespacing*{\section}{0pt}{2pt}{0pt}
\titlespacing*{\subsection}{0pt}{2pt}{0pt}
\renewcommand{\thesubsubsection}{\arabic{subsubsection}}
\titleformat{\subsubsection}[runin]{\itshape}{\thesubsubsection)}{1em}{}[:]
\titlespacing*{\subsubsection}{\parindent}{0pt}{*1}
\let\labelindent\relax
\usepackage{enumitem}
\setlist{nosep, before={\vspace{0.01\baselineskip}}, after={\vspace{0.01\baselineskip}}}

\usepackage{caption}
\usepackage{subcaption}
\usepackage{times}
\usepackage{amsmath}
\usepackage{amssymb}
\usepackage{booktabs}
\usepackage{array}
\usepackage{multirow}
\usepackage{threeparttable}

\author{Michael D. Friske}
\title{\LARGE \bf
	Integration of Augmented Reality and \\Mobile Robot Indoor SLAM for Enhanced Spatial Awareness
}

\begin{document}

\maketitle

\textbf{\textit{Abstract}---This research explores the integration of indoor Simultaneous Localization and Mapping (SLAM) with Augmented Reality (AR) to enhance situational awareness, improving safety in hazardous or emergency situations. The main contribution of this work is to enable mobile robots to provide real-time spatial perception to users who are not co-located with the robot. This is a comprehensive approach, including selecting suitable sensors for indoor SLAM, designing and building a platform, developing methods to display maps on AR devices, implementing this into software on an AR device, and improving the robustness of communication and localization between the robot and AR device in real-world testing. By taking this approach and analyzing each component of the integrated system, this paper highlights numerous areas for future research that can further advance the integration of SLAM and AR technologies. These advancements aim to significantly improve safety and efficiency during rescue operations.}
\section{Introduction}
\label{chapter:introduction}

In emergency situations, ensuring the safety of first responders and rescuers in unstable or hazardous environments is extremely challenging. Navigating through narrow spaces and locating survivors is a crucial task, but it can be hazardous for rescue teams. Robots are an essential tool that allows rescuers to traverse these dangerous areas with minimal risk. However, understanding the location of these robots relative to the rescue team in order to coordinate interventions can be challenging when using cameras on the robot in a first-person perspective.

Integrating mobile robot Simultaneous Localization and Mapping (SLAM) with Augmented Reality (AR) improves situational awareness by providing real-time capability to "see through walls" and enhance safety, efficiency, and accuracy of interventions - both in these emergency scenarios and non-emergency situations such as inspections or surveys.

Existing technologies like ground-penetrating radar and wall-penetrating radar such as the Xaver 1000\autocite{Camero2024} have limitations, such as limited penetration depth and resolution. This research focuses on designing a custom mobile robot with SLAM capabilities and developing an AR application to display its maps, offering a complementary approach to these existing technologies.

\vspace*{-0.08cm}
\section{Objectives}

The aim of this paper is to investigate how the integration of SLAM maps and an AR device can enhance situational awareness. It consists of four major objectives:

\textbf{1. Supporting Research:} 
A review of sensors and their compatibility with SLAM algorithms in indoor environments. Analysis of methods to estimate transformations between the robot and AR device for localization and mapping.

\textbf{2. Design and Construction of Mobile Robot:} 
Design and construction of a prototype mobile robot platform equipped with SLAM capabilities that generates maps to be used by the AR application.

\textbf{3. Development of AR Application:} 
Development of a prototype AR application running on an Android smartphone for displaying SLAM maps in an intuitive manner and maintaining localization between the device and the robot.

\textbf{4. Implementation and Integration:} 
Real-world testing and integration of the system to ensure reliable data transmission between the elements, accurate transformations and localization, and experimentation with visualization types.

\section{Sensors}
\label{chapter:sensors}

Sensors considered for the robot include Sound Detection and Ranging (Sonar) sensors\autocite{Borenstein1996, Chong2015}, Radio Detection and ranging (Radar) sensors\autocite{TexasInstruments2024, Sie2023}, Light Detection and Ranging (LiDAR)  sensors\autocite{Borenstein1996, Chong2015, Kucsera2006, Wang2021}, cameras \autocite{Borenstein1996, Chong2015, Chen2022}, and Inertial Measurement Units\autocite{Borenstein1996}.

\vspace*{-0.05cm}
\begin{figure}[H]
	\centering
	\includegraphics[width=0.9\linewidth]{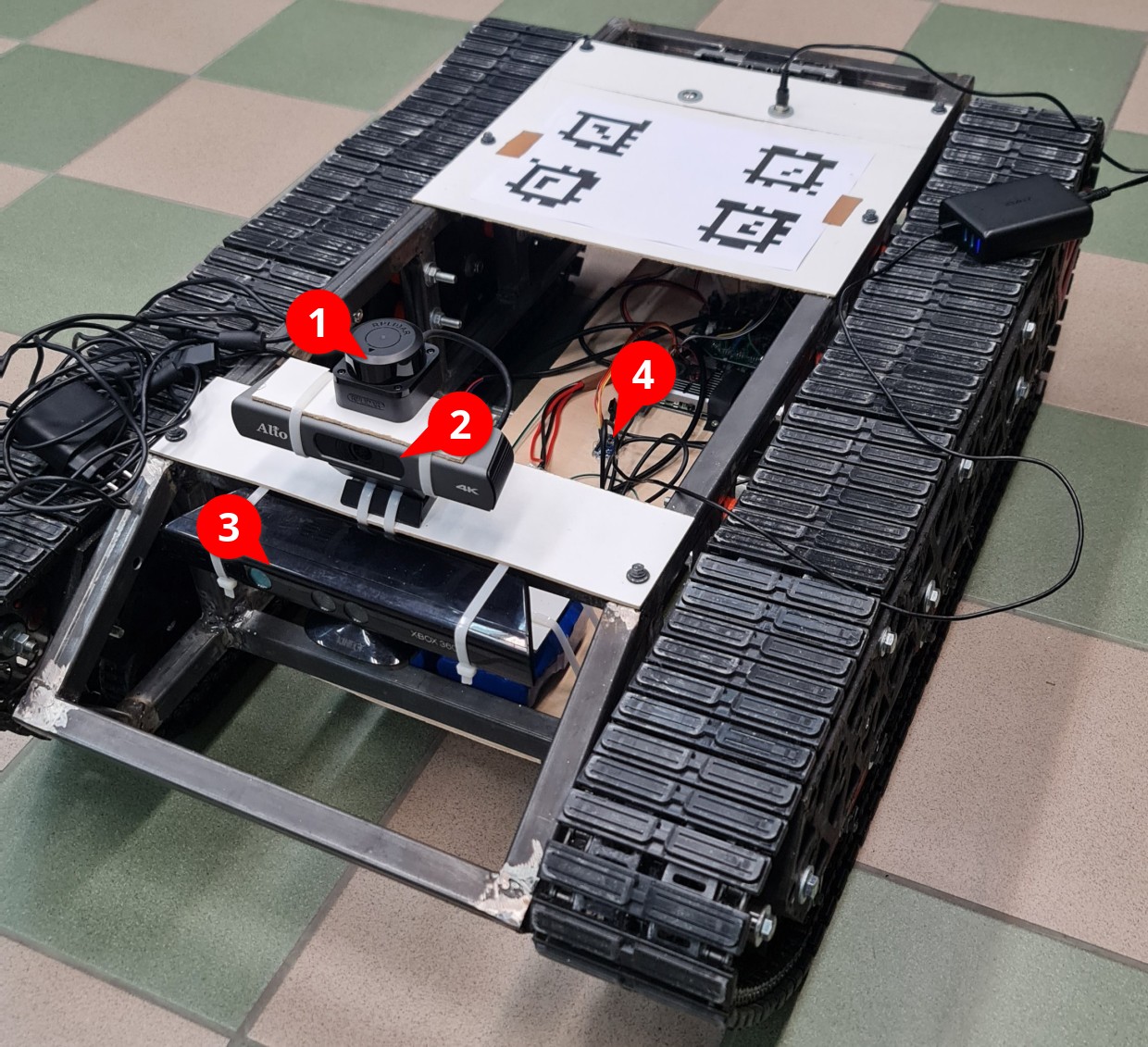}
	\caption{Sensors on the Robot Tank}
	\label{fig:final_assembled_labeled}
\end{figure}
\vspace*{-0.5cm}

The chosen sensors for the robot are: (1) 2D LiDAR, (2) monocular camera, (3) RGB-D camera, and (4) IMU (Shown in Figure \ref{fig:final_assembled_labeled}). These sensors were chosen based on their applicability to performing SLAM in indoor environments.

All sensors were calibrated to find their intrinsic and extrinsic properties\autocite{OpenCV2024, Brown1965}. The cameras required calibration with a known pattern\autocite{OpenCV2024,ROSPerception2020}. For the RGB-D cameras, calibration also required aligning depth and color images\autocite{OpenKinect2021, Wiedemeyer20142015}. IMU calibration used Allan variance analysis\autocite{Pupo2016, Woodman2007, ODRSG2022}. Kalibr was used to find extrinsics between the IMU and cameras\autocite{ASL2024, Furgale2013, Furgale2012}.

\section{Simultaneous Localization and Mapping}
\label{chapter:algorithms}

The data from sensors is utilized by SLAM algorithms to create maps for the AR device. To do so, the robot must localize itself while performing odometry and mapping the environment. Four categories of SLAM were explored:

\textbf{Visual SLAM}: Uses cameras to localize the robot and generate maps, and excels in environments with many visual features and sufficient lighting conditions \autocite{Chen2022, Zou2019, Theodorou2022}. Open source implementations considered include ORB-SLAM3, OpenVINS, PTAM, and CoSLAM\autocite{IntRoLab2021}.

\textbf{LiDAR SLAM}: Uses LiDAR data instead of cameras, unaffected by lighting\autocite{Chen2022, Zou2019, Theodorou2022}. Open source implementations considered include Hector SLAM, LeGO-LOAM, and LIO-SAM\autocite{IntRoLab2021}.

\textbf{Multi-sensor Fusion}: Couples LiDAR, visual sensors and/or IMUs to improve SLAM performance\autocite{Chen2022}. The open source implementation considered was RTAB-Map\autocite{IntRoLab2021}.

\textbf{Collaborative SLAM}: This extends any of the above categories to multiple agents, and can be implemented in a centralized or decentralized manner\autocite{Zou2019}.

Two SLAM algorithm implementations were chosen for the study:

\textbf{ORB-SLAM3}\autocite{Campos2020}: is a real-time SLAM library capable of various configurations, and was used with the monocular camera and IMU for visual and visual-inertial SLAM\autocite{Bowen2022, Jung2023}.

\textbf{RTAB-Map}\autocite{Labbe2024, IntRoLab2024}: is a comprehensive library built on a graph-based SLAM algorithm, generating dense point cloud maps for visualization. Visual odometry was used\autocite{Labbe2024}.
\section{Mobile Robot}\label{chapter:mobile_robot}

A SLAM system requires a robot capable of collecting environmental data. While Commercial Off-the-Shelf (COTS) robots or a robot borrowed from the University are options, designing and building a custom robot offers advantages such as greater flexibility in adapting the robot to suit specific requirements and bridging the gap between theory and practice. 

During the initial design phase, several robot platform archetypes were considered, including two-wheeled and four-wheeled holonomic and non-holonomic platforms, drones, and passive sensor systems. Ultimately, a robot tank was chosen for its simple differential drive system, small turn radius, and stability over small obstacles. Initial design constraints included a length of approximately one meter, a width of less than 80 cm, and a ground clearance of at least 15 cm with suspension travel of at least 10 cm.

\begin{figure}[H]
	\centering
	\includegraphics[width=0.9\linewidth]{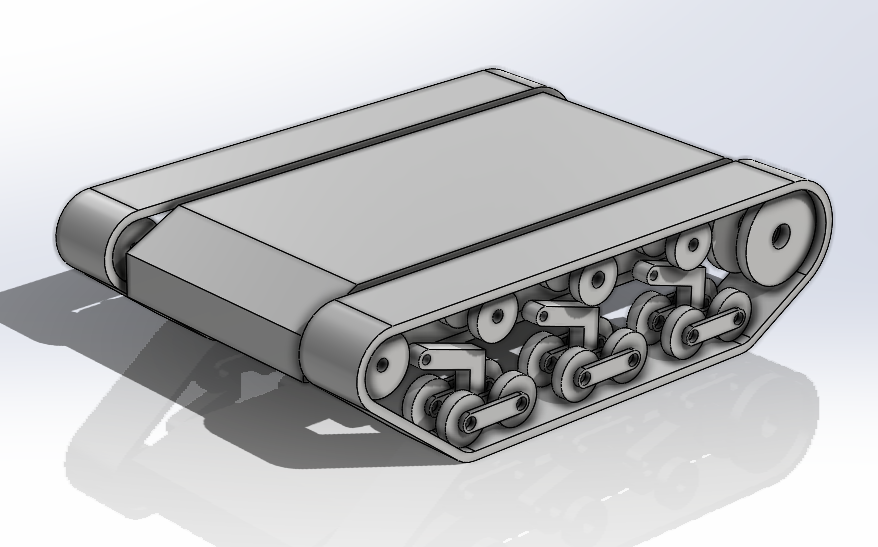}
	\caption{Robot Block Diagram}
	\label{fig:blockdiagram}
\end{figure}

\begin{figure}[H]
	\centering
	\includegraphics[width=0.9\linewidth]{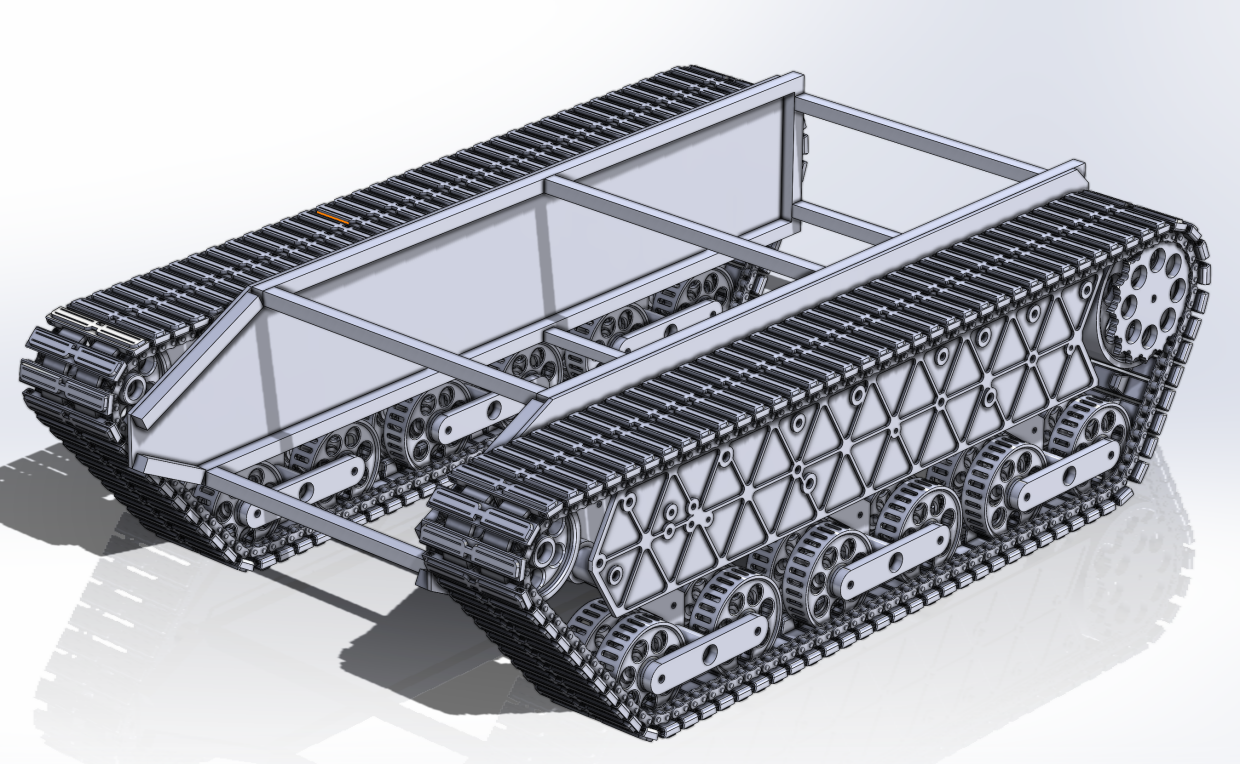}
	\caption{Final Design}
	\label{fig:isometricfinalbetter}
\end{figure}

An initial 2D outline formed the basis for the robot platform, which evolved through several iterations into the rough block diagram in Figure \ref{fig:blockdiagram} and ultimately to the final design in Figure \ref{fig:isometricfinalbetter}. 

\vspace*{-0.05cm}
\vspace*{-0.5cm}
\vspace*{-0.5cm}
Tube steel was welded to form the inner frame. Custom designed and wound springs were utilized for the suspension system, along with specialized track feet designed for this robot. The tracks use twelve roadwheels, six idler wheels, one tensioner with two additional idler wheels, and two custom drive sprockets per side for the track drive system. The lightweight wheels were iteratively designed to use minimal resin while maintaining strength, and repurposed hoverboard motors directly coupled to the drive sprockets provided adequate torque for the tracks\autocite{Manta2024}, eliminating the need for a gearbox. A coupler was designed from a 3D-scanned model of the motor hub and mated with custom keyed steel axles to transfer force between the motor and drive sprockets with minimal backlash. An external support bracket was added to stabilize the sprocket wheel.

Components were created using two techniques: fused deposition modeling (FDM) for strength in specific directions and stereolithography (SLA) for parts requiring isotropic properties such as wheels and tracks. Computer-aided Machining (CAM) in Fusion 360 was used to generate the instructions for computer-numerical control (CNC) machining of side panels and axles with keyways.

Electrical components like batteries, motor controllers, and wiring were mounted after assembly. Testing was conducted on the suspension, drivetrain, power supply, and sensors before integrating additional sensing devices.

\section{Software}
\label{chapter:software}

\subsection{General Architecture}
\begin{figure}[H]
	\centering
	\includegraphics[width=0.9\linewidth]{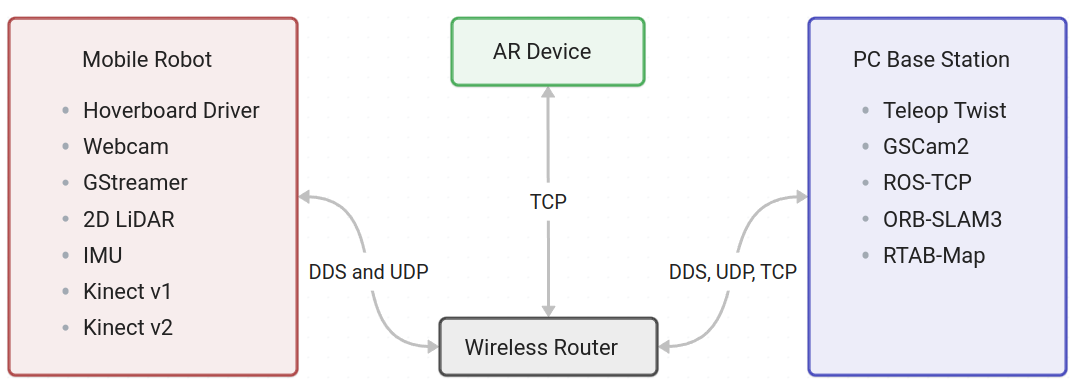}
	\caption{Overall Architecture}
	\label{fig:applicationarch}
\end{figure}

The system is divided into three major elements, shown in Figure \ref{fig:applicationarch}: the robot tank with a Raspberry Pi 4B, the base station PC, and the AR device. The Raspberry Pi 4B handles the hoverboard driver and sensors on the mobile robot. The base station PC runs the control software for the robot, processes sensor data, and performs the calculations for SLAM. The AR device is detailed separately in Section \ref{chapter:ar_application}. The system employs robot Operating System 2 (ROS2) as its framework, enabling simplified communication between each of the major elements and allowing for modular component-based design of the system.

\vspace*{-0.5cm}

\subsection{Mobile Robot Architecture}
The mobile robot uses a range of software components and libraries, with the architecture shown in Figure \ref{fig:robotarch}, each one of them is briefly described below:

\begin{figure}[H]
	\centering
	\includegraphics[width=0.9\linewidth]{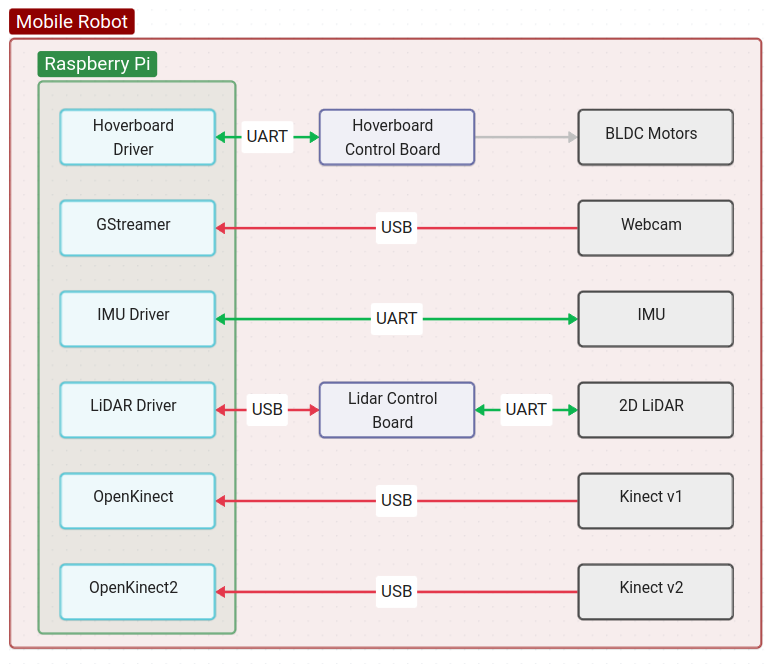}
	\caption{Mobile Robot Architecture}
	\label{fig:robotarch}
\end{figure}
\vspace*{-0.5cm}

\textbf{Hoverboard Driver}: An adapted version of an existing firmware and ROS2 driver for the hoverboard controller, operating in Field-Oriented Control Torque mode for accurate rotational control\autocite{Makarov2021, Feru2023}.

\textbf{Webcam}: A ROS2 package\autocite{Pitzer2023, ROSPerception2024} was initially used, but GStreamer\autocite{GStreamer2024} streaming MJPEG directly to the base station PC over User Datagram Protocol (UDP) improved performance.

\textbf{IMU}: The ST LSM9DS1 IMU was used, which contains a 3D accelerometer, gyroscope, and magnetometer. A Madgwick filter implementation was added for orientation estimation\autocite{Moeller2023, Garcia2024}.

\textbf{2D LiDAR}: The RPLiDAR C1, a low-cost 2D Direct Time-of-Flight laser range scanner was used with its purpose-built library\autocite{Slamtec2024}.

\textbf{Kinect v1}: The Kinect v1 is capable of outputting color and depth images as an RGB-D camera, and used the Freenect firmware and a ROS2 library\autocite{OpenKinect2022, Lio2023}, and worked well with the Raspberry Pi.

\textbf{Kinect v2}: The Kinect v2, using a time-of-flight camera sensor, provides more uniform point clouds but requires higher computational load and bandwidth than the Raspberry Pi was capable of and required the PC base station to be mounted to the robot\autocite{OpenKinect2021, Wiedemeyer20142015, Haaaaofei2023}.

\subsection{PC Base Station Architecture}

\begin{figure}[H]
	\centering
	\includegraphics[width=0.9\linewidth]{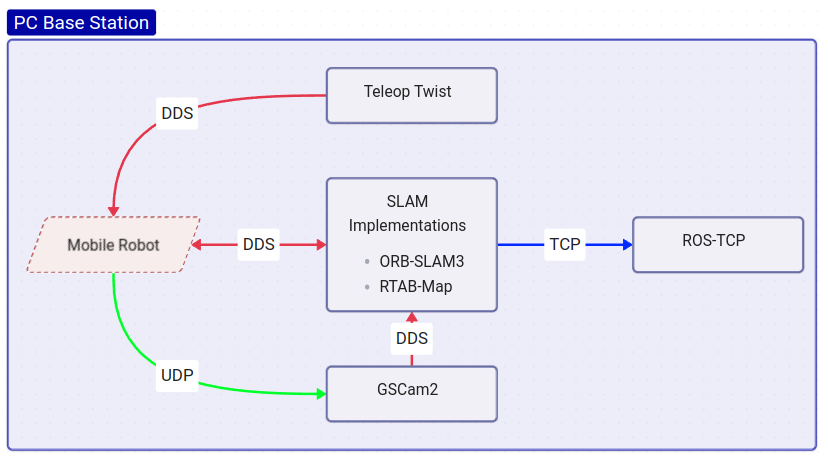}
	\caption{PC Base Station Architecture}
	\label{fig:pcarch}
\end{figure}

Each element of the PC base station architecture, shown in Figure \ref{fig:pcarch} and described below, receives data from the mobile robot, processes the data, and sends it to the AR application.

\vspace*{-0.5cm}

\textbf{Teleop Twist}:
Teleop twist, a built-in ROS2 package, allows sending manual control commands to the Raspberry Pi 4B and are interpreted by the hoverboard driver to drive the robot accordingly.

\textbf{GSCAM2}:
GStreamer on the Raspberry Pi 4B streams video to the base station PC, where GSCAM2\autocite{McQueen2024} captures the UDP stream, decodes it, and publishes it as ROS2 messages with timestamps encoded into the UDP stream.

\textbf{ROS TCP}:
A communication bridge between the PC Base Station and the AR Application, the ROS TCP package\autocite{UnityTechnologies2022} allows bi-directional communication between the two major elements.

\section{AR Application}
\label{chapter:ar_application}

The design of the AR application and device started with choosing the platform and framework. This application runs on a Samsung Note 20 Ultra. Initially, Android Studio, Unreal Engine, and Unity Engine were considered as application frameworks. After experimentation, Unity Engine was chosen for its balance of capability and ease of development.

\begin{figure}[H]
	\centering
	\includegraphics[width=0.9\linewidth]{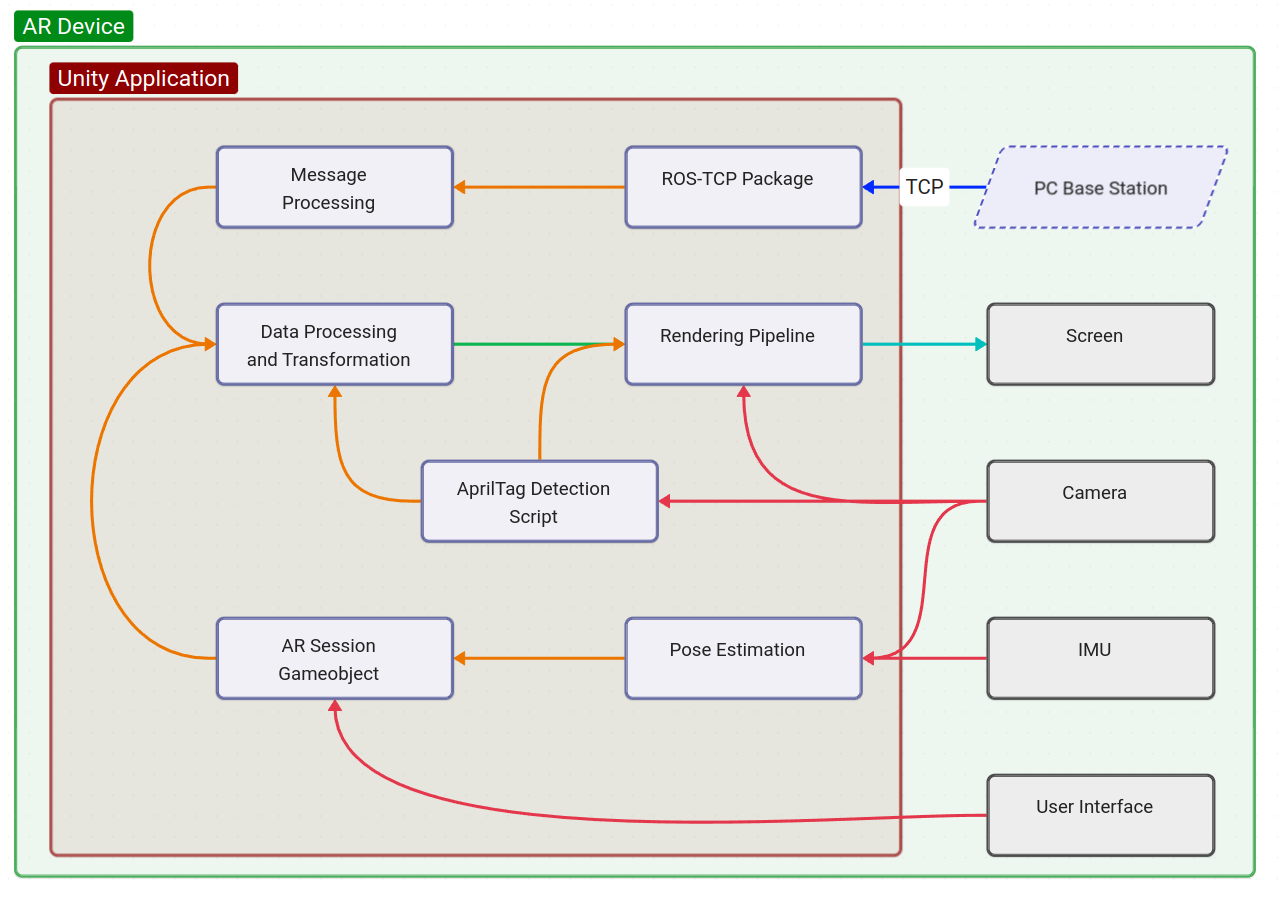}
	\caption{AR Application Architecture}
	\label{fig:ararch}
\end{figure}

The architecture of the application, shown in Figure \ref{fig:ararch}, forms the basis of the system. Each section below describes the components of the data flowchart in detail.

\vspace*{-0.5cm}
\textbf{Communication with PC Base Station}:
The mobile robot sends sensor messages to the PC base station, which are then converted into point cloud maps and sent to the AR device.

\textbf{Mobile Robot Pose Initialization}: 
The mobile robot's pose is initialized using AprilTags\autocite{Olson2011} mounted on the robot. The AR device attempts to locate AprilTags in its field of view, and set the global zero point at the robot's origin based on the known relative tag location.

\textbf{AR Device Pose Estimation}:
The Google ARCore library\autocite{Google2023} captures the camera image and renders it to the background of the application directly from the graphics processing unit (GPU). It also estimates pose transformation over time using IMU input fused with camera movement, correcting IMU drift.

\textbf{Data Processing and Transformation}:
The position data and the quaternion holding the zeroPointTransform object's rotation and the desired point size are stored in a 4x4 matrix called a Translation Rotation Scaling (TRS) matrix. 

The TRS matrix holding the zeroPointTransform is then transformed to the Unity world coordinate system. The point cloud data is then multiplied by this matrix, setting its positional and rotational origin to match the zeroPointTransform object, and stored temporarily in a buffer object. The results of all points are stored in an instancing matrix for rendering.

\textbf{Rendering Pipeline}:
The GPU performs a series of custom-written rendering steps to visualize the point cloud data. In the vertex shader stage, each point is assigned an instance ID, and its transformation in world space and clip space is determined. The distance between the camera and each point is computed, and a normalized vector representing a light source direction is transformed to world space coordinates\autocite{Nvidia2024}. The color of each point is retrieved using the instance ID. In the fragment shader stage, each point is colored based on the selected mode, chosen in the user interface. Five different modes were written for the fragment shader.

Unfortunately, with a high number of points, performance suffers. The revised approach implemented involved defining a single mesh whose topology consists of points, using a 32-bit vertex position buffer. This mesh is passed to the GPU using the DrawMesh method. In the vertex shader stage, points are drawn as true "point" objects, with their size in pixels adjusted dynamically based on the depth transformation to simulate perspective. The fragment shader stage processes the points based on the selected mode as before, and the lighting calculations were removed.
\section{Experimentation}	
\label{chapter:experimentation}

The experiments conducted aimed to enhance the overall system performance and were focused on several of the key components. Four experiments were conducted. The first experiment focused on improving the webcam and streaming performance to enhance the data acquired by the robot and the fidelity of the point cloud. The second experiment involved the performance of the AR application itself. The third experiment tested whether averaging the position of AprilTags improves localization accuracy, and the final experiment compared various map display methods.

\subsection{Webcam and Streaming Performance}

The goal of this first experiment was to improve the performance of webcam streaming from the mobile robot to the base station by reducing latency and improving the data rate to allow for higher resolutions and frame rates. This improved the inputs into the SLAM algorithms in order to enhance the accuracy of the resulting point clouds and the robot's localization within it.

\textbf{Setup:}
Several configurations were tested:
\begin{enumerate} 
	\item The initial setup used an unbranded USB webcam with a 65-degree field of view lens,
	\item The lens was replaced with a 120-degree field of view lens,
	\item The image transport library\autocite{ROSPerception2024} was used to compress the image,
	\item Raw MJPEG encoded in hardware on the webcam was forwarded as ROS2 messages,
	\item The unbranded webcam was replaced with a higher-performance webcam,
	\item GStreamer\autocite{GStreamer2024} and GSCAM\autocite{McQueen2024} with MJPEG to stream over UDP was used,
	\item Within GStreamer, H.264 encoding was applied instead of MJPEG. 
\end{enumerate}
\textbf{Results:} The generic USB webcam operated at 18 Hz at 640x480 on the Raspberry Pi but dropped to 7 Hz after transmission to the PC base station. Using GStreamer allowed for 1920x1080 streaming at 30 Hz, and H.264 encoding enabled 3840x2160 streaming at 30 Hz, though with compression artifacts that were not compatible with SLAM.

\textbf{Discussion and Conclusion:}
The choice of webcam had a significant impact on subjective image quality, but the streaming performance was primarily determined by the transmission protocol. GStreamer with UDP transport provided superior performance. While Test 7 achieved the highest resolution, the frame rate was identical, and the compression artifacts would be detrimental for SLAM algorithms. Therefore, Test 6 provided the most useful performance.

\vspace*{-0.4cm}
\begin{figure}[H]
	\centering
	\begin{minipage}[b]{0.49\linewidth}
		\resizebox{\columnwidth}{!}{%
			\begin{tikzpicture}
				\begin{axis}[
					xlabel={Test Number},
					ylabel={Frames Per Second (FPS)},
					symbolic x coords={1, 2, 3, 4, 5, 6, 7},
					xtick=data,
					ymin=0, ymax=35,
					ytick={0,5,10,15,20,25,30,35},
					ymajorgrids=true,
					grid style=dashed,
					]
					\addplot[
					color=red,
					mark=square,
					]
					coordinates {
						(1,7)(2,7)(3,6)(4,8)(5,7)(6,30)(7,30)
					};
				\end{axis}
			\end{tikzpicture}%
		}
		\caption{Webcam Frame rate}
	\end{minipage}
	\hfill
	\begin{minipage}[b]{0.49\linewidth}
		\resizebox{\columnwidth}{!}{%
			\begin{tikzpicture}
				\begin{axis}[
					xlabel={Test Number},
					ylabel={Latency (s)},
					symbolic x coords={1, 2, 3, 4, 5, 6, 7},
					xtick=data,
					ymin=0, ymax=5,
					ytick={0,1,2,3,4,5},
					ymajorgrids=true,
					grid style=dashed,
					]
					\addplot[
					color=red,
					mark=square,
					]
					coordinates {
						(1,2.95)(2,2.95)(3,4.85)(4,3.65)(5,3.55)(6,0.65)(7,0.75)
					};
				\end{axis}
			\end{tikzpicture}%
		}
		\caption{Webcam Latency}
	\end{minipage}
\end{figure}
\vspace*{-0.5cm}

\subsection{AR Application Performance}

The goal of this experiment was to enhance the performance of the AR application rendering large and dense point cloud maps by optimizing the data transformation and rendering processes, as mentioned in Section \ref{chapter:ar_application}.

\textbf{Setup:}
A point cloud containing 100,000 points of random position and colors within a 0.5 m cube was used to evaluate the performance of different rendering pipelines. The frame rate was logged and the median framerate after 30 seconds was used as the result. Four configurations were tested:

\begin{enumerate} 
	\item Processing and transformation of point cloud data on CPU, with individual points written to the rendering pipeline,
	\item Points written in batches to the rendering pipeline using the RenderMeshInstanced method,
	\item Processing on GPU using a compute shader, with point geometry written in parallel from GPU using a Shader Storage Buffer Object (SSBO),
	\item Processing on GPU in vertex shader, with point geometry written in a single batch using DrawMesh method. 
	\end{enumerate}
\vspace*{-0.5cm}
\textbf{Results:}
Test 4, which involved sending the data from the CPU to the GPU in a single batch in the AR application showed the best performance with a stable frame rate. In contrast, Test 3 failed to render the point cloud, due to specific hardware limitations of the mobile device used. Tests 1 and 2 were significantly slower.

\textbf{Discussion and Conclusion:}

The experiment revealed that the time taken to process and transform the point cloud on the CPU and to transmit the data to the GPU significantly impacts performance. Sending the data in a single batch, as in Test 4, significantly outperformed sending the data in multiple batches, as in Test 2. Minimizing the number of CPU-to-GPU data transfers is crucial for performance.

\subsection{AprilTag Position Averaging}

The goal of this experiment was to determine whether averaging the position of AprilTags over time improves the localization accuracy, both for position and rotation. 

\textbf{Setup:}
The experiment involved four test configurations, two with the device held in a static stand, and two with the device held by hand. Data was collected for 60 seconds at 30 Hz:

\begin{enumerate}
	\item Samples taken from 0.25 m above the AprilTags with the phone on a stand,
	\item Samples taken from 1.25 m at an angle with the phone on a stand,
	\item Samples taken holding the phone by hand from 0.25 m above the AprilTags,
	\item Samples taken holding the phone by hand from 1.25 m at an angle.
\end{enumerate}

\textbf{Results:}
Static tests showed low noise and high accuracy, while handheld tests, particularly at the larger distance, showed increased standard deviation. Statistical analysis techniques were applied to the collected data.

\textbf{Discussion and Conclusion:}
The statistical analysis showed that averaging frames together, counter-intuitively, increased the positional and rotational error of the detected AprilTag location when the phone was handheld. For best accuracy in handheld operations, a single still frame should be used. Averaging frames is still beneficial when the phone is mounted on a stable stand.

\subsection{Displaying AR Maps}

The goal of this experiment was to integrate the system as a whole and visualize the point cloud data to determine which visualization offered the best understanding of the environment.

\textbf{Setup:}
Both ORB-SLAM3 and RTAB-Map SLAM implementations were used with varying parameters to collect data for the point cloud, and five visualization methods were tested:

\begin{enumerate} 
	\item Distance-based Color,
	\item Axis-based Color,
	\item Depth Rainbow,
	\item Natural Color,
	\item Sonar Effect.
	\end{enumerate}

\textbf{Discussion and Conclusion:}
ORB-SLAM3 expects images to be black and white, resulting in a point cloud without color data. Combined with the challenge that a monocular map has no consistent scaling between initializations, requiring manually setting a scale factor for each test, the results of ORB-SLAM3 were not useful for visualization. However, RTAB-Map was extraordinarily effective. Although using the RGB-D camera as an input into ORB-SLAM3 may improve its results, RTAB-Map's point cloud density was far superior for visualization.

\begin{figure}[H]
	\centering
	\includegraphics[width=0.9\linewidth]{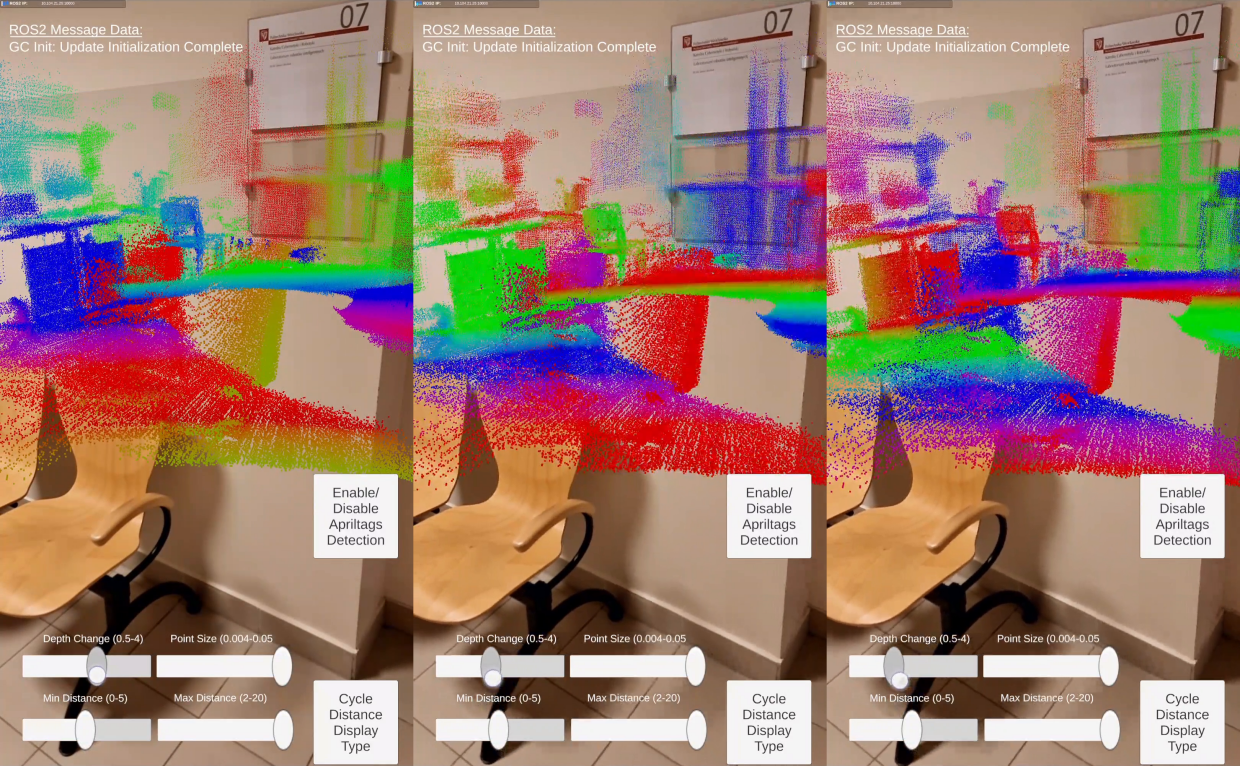}
	\caption{Depth Rainbow Visualization (Frequency between 1.2 m and 2.5 m) - RTAB-Map.}
	\label{fig:depthstoryboard}
\end{figure}

\begin{figure}[H]
	\centering
	\includegraphics[width=0.9\linewidth]{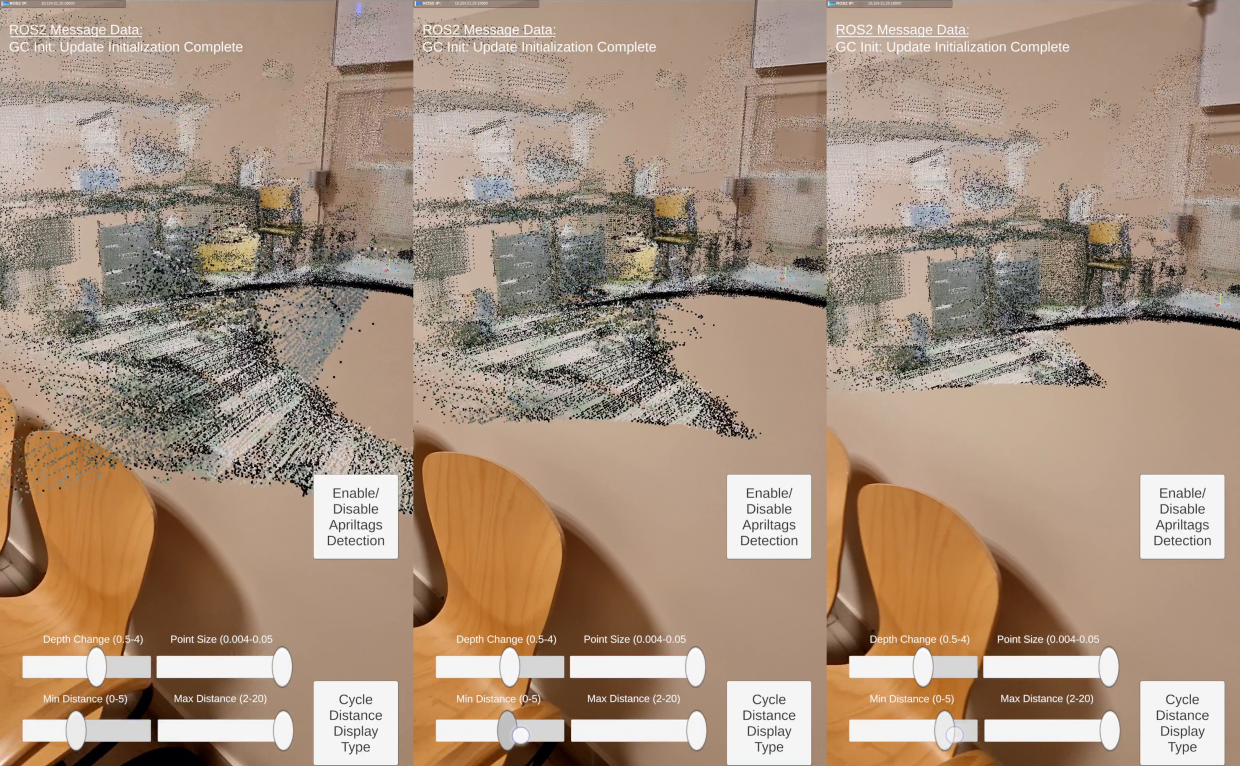}
	\caption{Natural Color Visualization (Cutoff between 2 m and 4 m) - RTAB-Map.}
	\label{fig:colorstoryboard}
\end{figure}

\begin{figure}[H]
	\centering
	\includegraphics[width=0.9\linewidth]{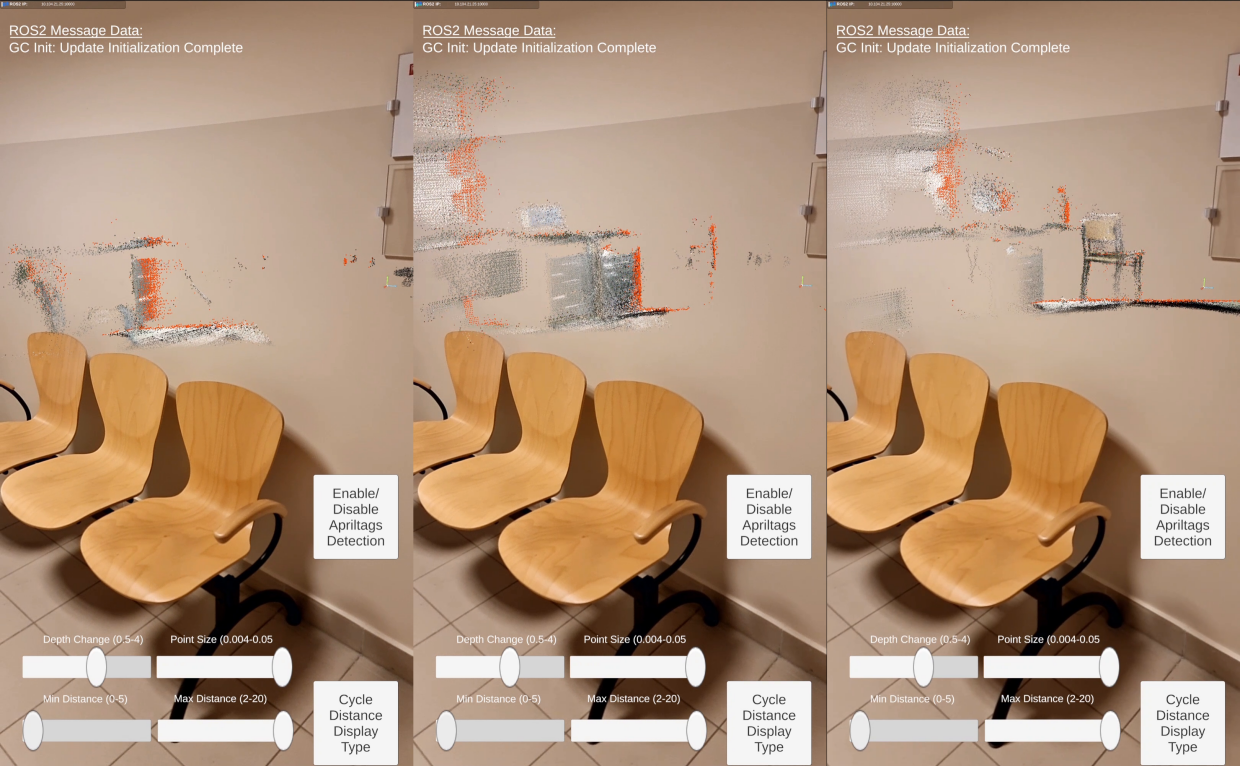}
	\caption{Sonar Visualization (Over 1 s) - RTAB-Map.}
	\label{fig:sonarstoryboard}
\end{figure}

The depth-rainbow visualization (Figure \ref{fig:depthstoryboard}) was most useful for understanding spatial relationships within the scene. The natural color visualization (Figure \ref{fig:colorstoryboard}) offered the best understanding of what each object in the scene was, but the colors were harder to identify. The "sonar" visualization (Figure \ref{fig:sonarstoryboard}) was the most useful for understanding depth. Moving the device to introduce parallax in the scene improved understanding of both the objects in the room and their spatial relationships.

\vspace*{-0.5cm}
\vspace*{-0.5cm}
\vspace*{-0.5cm}

\section{Discussion}
\label{discussion}

The entire process, from research to design to experimentation, provided key insights relating to the utility of SLAM mapping for enhancing situational awareness. ORB-SLAM3, despite its effectiveness in various SLAM applications, has limitations in providing human-intuitive visualization. The features it extracts lack color data and the point cloud is not dense enough to be intuitively understandable in AR. Conversely, RTAB-Map performed exceptionally well, with point cloud visualization significantly enhancing intuitive understanding of a room and increasing situational awareness by allowing AR users to see through walls.

The research foundation, mobile robot design, AR application development, and system integration were successful. Experimentation demonstrated that visualizing map data from the user's perspective enhances situational awareness. The transformation and localization between the robot and the AR device, performed only in an initialization step, were sufficient for accurate data display. The AR device effectively displayed transformed point cloud data in five visualizations, with the sonar visualization particularly effective.

\section{Future Work}

Several challenges and directions remain for future research: 
\vspace*{-0.5cm}
\begin{itemize} 
	\item Implementing other SLAM algorithms may offer better map visualizations,
	\item Robot autonomous navigation and AR-based goal setting within the AR application could further improve the system's utility for rescue operations,
	\item External localization and collaborative SLAM could improve overall accuracy and robustness,
	\item Referencing external AprilTags to define the world coordinate system could improve accuracy, 
	\item Point cloud filtering for clearer visualizations,
	\item Model fitting techniques such as RANSAC, applied to visualization and calibration, may open new visualization and calibration methods,
	\item Performing calculations onboard the mobile robot could reduce the need for high-bandwidth data transmission, improving system reliability,
	\item Integrating more or alternative sensors, such as a 3D LiDAR, could provide richer environmental data, enhancing the SLAM process and AR visualization. 
	\end{itemize}

\section{Conclusion}
\label{chapter:conclusion}

In emergency situations, ensuring the safety of first responders in hazardous environments while locating and rescuing survivors is challenging. This research has demonstrated that integrating mobile robot SLAM (Simultaneous Localization and Mapping) maps with Augmented Reality (AR) can significantly enhance situational awareness, offering a robust tool to visualize dangerous areas with minimal risk. By achieving the objectives set out in this paper, the system not only offers the ability to improve the safety and efficiency of rescue operations but also has potential applications in non-emergency scenarios, such as inspections or surveys.

The AR-enhanced visualization allows users to view SLAM-generated maps in real time from their perspective, effectively "seeing through walls" and better understanding the robot's environment relative to their own position. While this study offers a significant contribution to the field, particularly in complementing existing technologies like ground-penetrating and wall-penetrating radars, it also lays the groundwork for future research. There remains much to explore, including the performance of alternative SLAM algorithms, enhancement of visualization techniques, and integration of additional sensors which could further extend the system's capabilities and applicability.

This research underscores the potential of combining AR with mobile robot SLAM to improve situational awareness in critical applications, making it a promising area for continued investigation and research.

\renewcommand{\bibfont}{\footnotesize}
\printbibliography

@Article{Theodorou2022,
  author       = {Theodorou, Charalambos and Velisavljevic, Vladan and Dyo, Vladimir and Nonyelu, Fredi},
  date         = {2022-09},
  journaltitle = {Array},
  title        = {Visual SLAM algorithms and their application for AR, mapping, localization and wayfinding},
  doi          = {10.1016/j.array.2022.100222},
  issn         = {2590-0056},
  pages        = {100222},
  volume       = {15},
  comment      = {Three types of SLAM
- Monocular
- Stereo (everywhere)
- RGB-D (indoors)

Slam Algorithms
- MonoSLAM
- PTAM
- PTAM-Dense
- DTAM
- SLAM++

But these are all for motionless environments

SLAM algorithms have three basic modules
- Initialization (developing global coordinate system and initial reference region)
- Tracking
- Mapping

DOC-SLAM can remove moving objects from the SLAM data

ORB-SLAM3 outperforms everything else for monocular and stereo SLAM},
  journal      = {Array},
  month        = sep,
  publisher    = {Elsevier BV},
  readstatus   = {read},
  year         = {2022},
}

@Article{Zou2019,
  author       = {Zou, Danping and Tan, Ping and Yu, Wenxian},
  date         = {2019-10},
  journaltitle = {Virtual Reality and Intelligent Hardware},
  title        = {Collaborative visual SLAM for multiple agents:A brief survey},
  doi          = {10.1016/j.vrih.2019.09.002},
  issn         = {2096-5796},
  number       = {5},
  pages        = {461--482},
  volume       = {1},
  comment      = {Can be centralized or decentralized (centralized easier to implement)

- CoSLAM, PTAMM (extension of PTAM), C2TAM, MOARSLAM, CCM-SLAM, CVI-SLAM
- Issue with this is that more cameras = more throughput needed
- Can do some amount of pre-processing on the system itself, rather than relying on the 'base station' to do all of the processing and taking in all the live video etc
- Can even do state estimation, control, planning on each robot itself and then send to centralized server for fuse maps into global map
- Using EKF is perhaps better for some things, but a key-frame approach is easier to extend to multiple agents},
  journal      = {Virtual Reality and Intelligent Hardware},
  month        = oct,
  publisher    = {Elsevier BV},
  readstatus   = {read},
  year         = {2019},
}

@Article{Campos2020,
  author     = {Campos, Carlos and Elvira, Richard and Rodríguez, Juan J. Gómez and Montiel, José M. M. and Tardós, Juan D.},
  date       = {2020},
  title      = {ORB-SLAM3: An Accurate Open-Source Library for Visual, Visual-Inertial and Multi-Map SLAM},
  doi        = {10.48550/ARXIV.2007.11898},
  copyright  = {arXiv.org perpetual, non-exclusive license},
  publisher  = {arXiv},
  readstatus = {read},
  year       = {2020},
}

@Article{Labbe2024,
  author     = {Labbé, Mathieu and Michaud, François},
  date       = {2024},
  title      = {RTAB-Map as an Open-Source Lidar and Visual SLAM Library for Large-Scale and Long-Term Online Operation},
  doi        = {10.48550/ARXIV.2403.06341},
  copyright  = {arXiv.org perpetual, non-exclusive license},
  publisher  = {arXiv},
  readstatus = {read},
  year       = {2024},
}

@Article{Chong2015,
  author       = {Chong, T.J. and Tang, X.J. and Leng, C.H. and Yogeswaran, M. and Ng, O.E. and Chong, Y.Z.},
  date         = {2015},
  journaltitle = {Procedia Computer Science},
  title        = {Sensor Technologies and Simultaneous Localization and Mapping (SLAM)},
  doi          = {10.1016/j.procs.2015.12.336},
  issn         = {1877-0509},
  pages        = {174--179},
  volume       = {76},
  journal      = {Procedia Computer Science},
  publisher    = {Elsevier BV},
  readstatus   = {read},
  year         = {2015},
}

@Article{Chen2022,
  author       = {Chen, Weifeng and Zhou, Chengjun and Shang, Guangtao and Wang, Xiyang and Li, Zhenxiong and Xu, Chonghui and Hu, Kai},
  date         = {2022-11},
  journaltitle = {Remote Sensing},
  title        = {SLAM Overview: From Single Sensor to Heterogeneous Fusion},
  doi          = {10.3390/rs14236033},
  issn         = {2072-4292},
  number       = {23},
  pages        = {6033},
  volume       = {14},
  day          = {28},
  journal      = {Remote Sensing},
  month        = {11},
  publisher    = {MDPI AG},
  readstatus   = {read},
  year         = {2022},
}

@Article{Kucsera2006,
  author     = {Kucsera, P{\'e}ter},
  title      = {Sensors for mobile robot systems},
  number     = {4},
  pages      = {645--658},
  volume     = {5},
  journal    = {Academic and Applied Research in Military Science},
  publisher  = {Citeseer},
  readstatus = {read},
  year       = {2006},
}

@Article{Borenstein1996,
  author     = {Borenstein, Johann and Everett, HR and Feng, Liqiang and others},
  title      = {Where am I? Sensors and methods for mobile robot positioning},
  number     = {120},
  pages      = {27},
  volume     = {119},
  journal    = {University of Michigan},
  publisher  = {Citeseer},
  readstatus = {read},
  year       = {1996},
}

@Software{Makarov2021,
  author     = {Alex Makarov},
  title      = {hoverboard-robotics/hoverboard-firmware-hack-FOC, forked from EFeru/hoverboard-firmware-hack-FOC},
  url        = {https://github.com/hoverboard-robotics/hoverboard-firmware-hack-FOC},
  year       = {2021},
  readstatus = {read},
}

@Software{Feru2023,
  author     = {Emanuel Feru},
  title      = {hoverboard-firmware-hack-FOC},
  url        = {https://github.com/EFeru/hoverboard-firmware-hack-FOC},
  year       = {2023},
  readstatus = {read},
}

@Software{Pitzer2023,
  author     = {Benjamin Pitzer},
  title      = {{usb-cam - A ROS Driver for V4L2-USB Cameras}},
  url        = {https://github.com/ros-drivers/usb_cam/tree/ros2},
  version    = {0.7.0},
  year       = {2023},
  readstatus = {read},
}

@Software{GStreamer2024,
  author     = {GStreamer},
  title      = {GStreamer open-source multimedia framework},
  url        = {https://github.com/GStreamer/gstreamer},
  version    = {1.24.0},
  year       = {2024},
  readstatus = {read},
}

@Software{McQueen2024,
  author     = {Clyde McQueen},
  title      = {GSCam2 - ROS2 camera driver for GStreamer-based video streams},
  url        = {https://github.com/clydemcqueen/gscam2},
  year       = {2024},
  readstatus = {read},
}

@Software{Moeller2023,
  author     = {Niels Moeller},
  title      = {LSM9DS1-Handler - A C++/Linux driver and ROS2 wrapper for the LSM9DS1 9DOF IMU},
  url        = {https://github.com/nm47/lsm9ds1_handler},
  year       = {2023},
  readstatus = {read},
}

@Software{Slamtec2024,
  author     = {Slamtec},
  title      = {RPILidar-ROS},
  url        = {https://github.com/Slamtec/rplidar_ros},
  version    = {2.15},
  year       = {2024},
  readstatus = {read},
}

@Software{OpenKinect2022,
  author     = {OpenKinect},
  title      = {libfreenect - Drivers and libraries for the Xbox Kinect device on Windows, Linux, and OS X},
  url        = {https://github.com/OpenKinect/libfreenect},
  version    = {0.6.4},
  year       = {2022},
  readstatus = {read},
}

@Software{OpenKinect2021,
  author     = {OpenKinect},
  title      = {libfreenect2 - Open source drivers for the Kinect for Windows v2 device},
  url        = {https://github.com/OpenKinect/libfreenect2},
  version    = {0.2.1},
  year       = {2021},
  readstatus = {read},
}

@Software{Lio2023,
  author     = {Fernando de Lio},
  title      = {Kinect-ROS2 - C++ ROS2 driver for Kinect v1 (Xbox 360).},
  url        = {https://github.com/fadlio/kinect_ros2},
  year       = {2023},
  readstatus = {read},
}

@Software{Haaaaofei2023,
  author     = {Haaaaofei},
  title      = {Kinect2-ROS2 - Tools for using the Kinect One (Kinect v2) in ROS2},
  url        = {https://github.com/YuLiHN/kinect2_ros2},
  year       = {2023},
  readstatus = {read},
}

@Software{Wiedemeyer20142015,
  author       = {Wiedemeyer, Thiemo},
  title        = {{IAI Kinect2}},
  url          = {https://github.com/code-iai/iai_kinect2},
  year         = {2014 -- 2015},
  organization = {Institute for Artificial Intelligence},
  address      = {University Bremen},
  readstatus   = {read},
}

@Software{ASL2024,
  author     = {{Autonomous Systems Lab}},
  title      = {Kalibr - The Kalibr visual-inertial calibration toolbox},
  url        = {https://github.com/ethz-asl/kalibr},
  year       = {2024},
  readstatus = {read},
}

@Software{ODRSG2022,
  author     = {{Russell Buchanan, Oxford Dynamic Robot Systems Group}},
  title      = {Allan-Variance-ROS - ROS compatible tool to generate Allan Deviation plots},
  url        = {https://github.com/ori-drs/allan_variance_ros},
  year       = {2022},
  readstatus = {read},
}

@Software{UnityTechnologies2022,
  author     = {Unity~Technologies},
  title      = {ROS-TCP-Endpoint - ROS package used to create an endpoint to accept ROS messages sent from a Unity scene using the ROS TCP Connector scripts},
  url        = {https://github.com/Unity-Technologies/ROS-TCP-Endpoint/tree/main-ros2},
  version    = {0.7.0},
  year       = {2022},
  readstatus = {read},
}

@Software{Bowen2022,
  author     = {Chen Bowen},
  title      = {ROS2-ORBSLAM3 - A primary, extremely simple package linking orbslam3 and ros2},
  url        = {https://github.com/curryc/ros2_orbslam3},
  year       = {2022},
  readstatus = {read},
}

@Software{Jung2023,
  author     = {Haebeom Jung},
  title      = {ORB-SLAM3-ROS2 - ROS2 wrapping package for orbslam3 library},
  url        = {https://github.com/zang09/ORB_SLAM3_ROS2/tree/humble},
  year       = {2023},
  readstatus = {read},
}

@Software{IntRoLab2024,
  author     = {IntRoLab},
  title      = {RTAB-Map - RTAB-Map library and standalone application},
  url        = {https://github.com/introlab/rtabmap},
  version    = {0.21.4},
  year       = {2024},
  readstatus = {read},
}

@WWW{IntRoLab2021,
  author     = {IntRoLab},
  date       = {2021-07},
  title      = {List of Open Source SLAM projects},
  url        = {https://github.com/introlab/rtabmap/wiki/List-of-Open-Source-SLAM-projects},
  readstatus = {read},
}

@Article{Furgale2013,
  author     = {Furgale, Paul and Rehder, Joern and Siegwart, Roland},
  date       = {2013-11},
  title      = {Unified Temporal and Spatial Calibration for Multi-sensor Systems},
  doi        = {10.1109/iros.2013.6696514},
  booktitle  = {In Proceedings of the IEEE/RSJ International Conference on Intelligent Robots and Systems (IROS), Tokyo, Japan},
  publisher  = {IEEE},
  readstatus = {read},
}

@InProceedings{Furgale2012,
  author     = {Furgale, Paul and Barfoot, Timothy D. and Sibley, Gabe},
  booktitle  = {2012 IEEE International Conference on Robotics and Automation},
  date       = {2012-05},
  title      = {Continuous-time batch estimation using temporal basis functions},
  doi        = {10.1109/icra.2012.6225005},
  location   = {St. Paul, MN},
  pages      = {2088--2095},
  publisher  = {IEEE},
  readstatus = {read},
}

@Misc{Pupo2016,
  author     = {Pupo, Leslie},
  title      = {Departament de Teoria del Senyal i Comunicacions},
  readstatus = {read},
  year       = {2016},
}

@WWW{Camero2024,
  author     = {Camero},
  title      = {Xaver 1000},
  url        = {https://camero-tech.com/xaver-products/xaver-1000/},
  readstatus = {read},
}

@Software{Google2023,
  author     = {Google},
  title      = {ARCore: Augmented Reality Platform for Android},
  url        = {https://developers.google.com/ar},
  year       = {2023},
  readstatus = {read},
}

@Software{ROSPerception2020,
  author     = {{ROS-Perception}},
  title      = {Image-Pipeline - An image processing pipeline for ROS.},
  url        = {https://github.com/ros-perception/image_pipeline},
  version    = {2.1.1},
  year       = {2020},
  readstatus = {read},
}

@WWW{Manta2024,
  author     = {Manta},
  date       = {2024},
  title      = {MSB9003 MAMBA Smart Balance Board 6.5’},
  url        = {https://manta.com.pl/produkt/msb9003/},
  readstatus = {read},
}

@Article{Wang2021,
  author     = {Wang, Han and Wang, Chen and Xie, Lihua},
  title      = {Lightweight 3-D Localization and Mapping for Solid-State LiDAR},
  doi        = {10.1109/lra.2021.3060392},
  issn       = {2377-3774},
  number     = {2},
  pages      = {1801–1807},
  volume     = {6},
  journal    = {IEEE Robotics and Automation Letters},
  month      = apr,
  publisher  = {Institute of Electrical and Electronics Engineers (IEEE)},
  readstatus = {read},
  year       = {2021},
}

@Article{Sie2023,
  author     = {Emerson Sie and Xinyu Wu and Heyu Guo and Deepak Vasisht},
  title      = {Radarize: Large-Scale Radar {SLAM} for Indoor Environments},
  doi        = {10.48550/ARXIV.2311.11260},
  eprint     = {2311.11260},
  eprinttype = {arXiv},
  url        = {https://doi.org/10.48550/arXiv.2311.11260},
  volume     = {abs/2311.11260},
  bibsource  = {dblp computer science bibliography, https://dblp.org},
  journal    = {CoRR},
  readstatus = {read},
  year       = {2023},
}

@WWW{TexasInstruments2024,
  author     = {{Texas Instruments}},
  date       = {2024},
  title      = {mmWave radar sensors},
  url        = {https://www.ti.com/sensors/mmwave-radar/overview.html},
  readstatus = {read},
}

@WWW{OpenCV2024,
  author     = {OpenCV},
  date       = {2024},
  title      = {OpenCV: Camera Calibration},
  url        = {https://docs.opencv.org/4.x/dc/dbb/tutorial_py_calibration.html},
  readstatus = {read},
}

@Misc{Brown1965,
  author     = {Duance C. Brown},
  title      = {Decentering Distortion of Lenses},
  readstatus = {read},
  year       = {1965},
}

@Software{ROSPerception2024,
  author     = {{ROS-Perception}},
  title      = {Image-Transport-Plugins - A set of plugins for publishing and subscribing to Image topics in representations other than raw pixel data.},
  url        = {https://github.com/ros-perception/image_transport_plugins},
  version    = {4.0.0},
  year       = {2024},
  readstatus = {read},
}

@InProceedings{Olson2011,
  author     = {Edwin Olson},
  booktitle  = {Proceedings of the {IEEE} International Conference on Robotics and Automation ({ICRA})},
  title      = {{AprilTag}: A robust and flexible visual fiducial system},
  pages      = {3400-3407},
  publisher  = {IEEE},
  month      = may,
  readstatus = {read},
  year       = {2011},
}

@WWW{Garcia2024,
  author     = {Mario García},
  date       = {2024},
  title      = {AHRS - Altitude and Heading Reference System},
  url        = {https://pypi.org/project/AHRS/},
  readstatus = {read},
}

@WWW{Nvidia2024,
  author     = {Nvidia},
  date       = {2024},
  title      = {The Cg Tutorial, Chapter 5. Lighting},
  url        = {https://developer.download.nvidia.com/CgTutorial/cg_tutorial_chapter05.html},
  readstatus = {read},
}

@TechReport{Woodman2007,
  author      = {Woodman, Oliver J.},
  institution = {University of Cambridge, Computer Laboratory},
  title       = {{An introduction to inertial navigation}},
  doi         = {10.48456/tr-696},
  number      = {UCAM-CL-TR-696},
  url         = {https://www.cl.cam.ac.uk/techreports/UCAM-CL-TR-696.pdf},
  month       = aug,
  readstatus  = {read},
  year        = {2007},
}

\end{document}